\pdfoutput=1
\documentclass[11pt]{article}

\usepackage[]{acl}

\usepackage{times}
\usepackage{latexsym}

\usepackage{algorithm}
\usepackage{algpseudocode}

\usepackage[T1]{fontenc}
\usepackage[utf8]{inputenc}

\usepackage{microtype}

\usepackage{amsmath, amssymb} % For better math
\newcommand{\vect}[1]{\boldsymbol{#1}}

\title{Automatic Generation of Factual News Headlines in Finnish}

\author{Maximilian Koppatz\textsuperscript{1}, Khalid Alnajjar\textsuperscript{1}, Mika Hämäläinen\textsuperscript{1} and Thierry Poibeau\textsuperscript{2} \\
  \textsuperscript{1} University of Helsinki \\
  \textsuperscript{2} École Normale Supérieure-PSL and CNRS and Université Sorbonne nouvelle \\
  \texttt{firstname.lastname@helsinki.fi} \\}

\begin{document}
\maketitle
\begin{abstract}
We present a novel approach to generating news headlines in Finnish for a given news story. We model this as a summarization task where a model is given a news article, and its task is to produce a concise headline describing the main topic of the article. Because there are no openly available GPT-2 models for Finnish, we will first build such a model using several corpora. The model is then fine-tuned for the headline generation task using a massive news corpus. The system is evaluated by 3 expert journalists working in a Finnish media house. The results showcase the usability of the presented approach as a headline suggestion tool to facilitate the news production process.
\end{abstract}

\section{Introduction}

Authoring a good headline is an essential step in the process of writing and publishing news articles. A good headline should be an apt and concise description of the contents of the article. It should also be captivating so that it makes a potential reader interested in reading the article in addition to following the guidelines set by the news agency in question. Good and bad headlines have also a great impact on the number of visitors \cite{headlines1,headlines2}, which directly translates into revenue on ad supported news websites.

It is very typical to use A/B testing to study which headline candidates are more successful in engaging users. This testing requires there to be headline candidates to test to begin with. For this reason, editors need to write multiple headline candidates for a single news article. This task takes a lot of time away from other editorial work especially since the people inventing the headlines are very often not the same people who write the articles. According to journalists working at the Finnish press house Sanoma, editors often times invent dozens of alternative headlines a day for news articles. Needless to say there is a commercial interest in automating this task.

It is not straightforward to automatically generate news headlines that are useful. A usable headline is expected to convey correct facts and be thematically relevant. At the same time, there must be diversity in the generated headlines given that the press houses want to have access to multiple headline variants. There is a communicative-creative trade off (see \citet{hamalainen-honkela-2019-co}) in how creative the system can be while still conveying the desired factual meaning. Additionally, the generated headlines must be interesting, because no reader would read a news story that sound boring from the very beginning.

In this paper, we represent a method for conditional headline generation by using a generative autoregressive Transformer \cite{transformer} based language model that is fine-tuned for the headline generation task in Finnish. The approach we follow can be seen as a special case of text summarization. Instead of the target being a full summary, it is a very compact headline. The reason for approaching the problem for this angle instead of resorting to a masked language model such as BERT~\cite{bert}, is that training autoregressive language models is computationally easier and faster than masked language models. This is because masked language models are trained to predict only a small percentage of words in a text during each forward pass, while autoregressive language models predict every word during every pass.

The main contributions of this paper are the following:

\begin{itemize}
  \item We train the first Finnish GPT-2~\cite{gpt2} model
  \item We fine-tune the model for the downstream task of headline generation for the morphologically rich Finnish language
  \item We present a human evaluation using real journalists who invent headlines as their day job
\end{itemize}

\section{Related Work}

In contemporaneous studies, neural headline generation is approached from the point of view of text summarization. Text summarization in itself has traditionally been divided into extractive and abstractive summarization. Currently, both of these types of tasks are tackled with approaches that utilize the Transformer \cite{copycopy,bukhtiyarov2020advances, liu2019text}. 

A common type of an approach for both extractive and abstractive summarization is an encoder-decoder type language model such as BertSumExt \cite{liu2019text} and PEGASUS \cite{zhang2020pegasus}. Summarization as a seq2seq problem suits the encoder-decoder model-paradigm well as you have a source and a target text like in NMT problems. In this setup, abstractive summarization is performed by the generative decoder part. For purely extractive tasks, the decoder is often replaced by some form of classifier which selects which tokens in the input should be in the resulting summary.

Another type of an approach, and the one used in this paper, is to fine-tune a GPT-2 \cite{gpt2} style auto-regressive language model for the summarization task \cite{kieuvongngam2020automatic,copycopy}. These approaches perform some form of concatenation of the target summary to the end of the source text with special tokens as delimiters between source and target. This approach does not fit the paradigm of sequence transduction as well as encoder-decoder setups, but does have its advantages. For one, all of the parameters are reused maximally, as the entire model network is pre-trained on text generation. When fine-tuning for the summary, this continues to be the case. Encoder-decoder setups tend to become more complex.

Recent headline generation approaches tend to use some form of BLEU or ROUGE for automatic evaluation of the models \cite{headlinegen1,bukhtiyarov2020advances,tilk-alumae-2017-low}. These metrics are naturally used for summarization as well. Furthermore, Beam Search is a common way to perform the generation.

For Finnish in particular, the literature for neural headline generation and summarization is scarce. Currently however, most work regarding Finnish headline and text generation seems to be using more conventional NLP, rule-based and statistical methods \cite{ journalisticnlg, hamalainen-alnajjar-2019-generating,cf07d21e4a91473b95b374ab67d2bab2}.

For news focused Finnish NLP there has been work on generating sports reports from event data using a pointer-generation network \cite{Kanerva2019TemplatefreeDG}, generation of creative headlines in Finnish using templates \cite{colorfulfinnishheadlines} and rumor detection in Finnish news \cite{1bd9ff9b61a24eea9e6b33bf13a5fc56} using BERT and LSTMs.

How creative NLG systems are evaluated has been investigated in recent work \cite{evaluation2021}, and the evaluation done in this paper roughly follows the conclusions drawn. Specifically, to evaluate the generation of the model aligned with what task the model was designed and trained to perform \cite{3a5a5943607346358b78a2cb12d49ee0}. Following this idea, the evaluation of the model in this paper is not relying simply on offline metrics, but on manual structured review by domain-experts with criteria relevant to the real-world use case.

\section{Data}

This section details the data, filtering, processing and tokenization used in this paper. There are two separate modeling tasks we perform: unsupervised generative pre-training and generative fine-tuning. For this reason, we make sure that there are always two columns in the dataset: "body" and "title". The body column contains everything except headlines and is used as the pre-training data. Later, the headlines are added for the fine-tuning task.

\subsection{Corpora}
The data used for pre-training the language model consists of four corpora concatenated together: Sanoma, Wikipedia, Yle and Ylilauta.

\textbf{The Sanoma corpus} is our primary and largest corpus. It is a proprietary corpus of news articles from the most important Finnish news paper \textit{Helsingin Sanomat} and the widely spread yellow press paper \textit{Ilta-Sanomat}. This corpus contains approximately 3.8 million Sanoma news articles published between the year 1990 and 2021. The topic coverage is as broad as one would expect from news media, ranging from domestic and international politics to sports and culture events. 

The Sanoma corpus contains the headline, ingress, and article body for most articles. We concatenate the ingress to the body text with double newlines between. This data was saved into parquet format with "title" and "body" columns. Headlines are kept separate because they are used only in the fine-tuning phase. This holds true to all corpora with headlines.

\textbf{Wikipedia} is a great corpus for language modeling, as it is freely available and contains information about the world. The corpus contains pages containing information about countries, people, history, science and much more. This is particularly useful for unsupervised language model pre-training as the model can learn from the information found in Wikipedia. The Finnish Wikipedia dump\footnote{https://dumps.wikimedia.org/fiwiki/latest/} from 24.11.2020 was used. This dump was parsed into a parquet file with again a "title" and "body" column. The dump contains 463,780 pages.

A corpus of news articles from \textbf{Yle}\footnote{http://urn.fi/urn:nbn:fi:lb-2017070501} was parsed into the same "title" and "body" format as the Sanoma and Wikipedia corpora. This corpus is small and only contributed around 100 000 articles. 

The \textbf{Ylilauta} corpus\footnote{http://urn.fi/urn:nbn:fi:lb-2016101210} contains 335,004 messages from the Ylilauta forums. These messages are quite different to the rest of the data used, as this is not structured text. This text is also colloquial. Furthermore, this corpus does not contain headlines. As it represents a different textual domain, it makes it possible for the model to learn a representation of colloquial Finnish as well.

\subsection{Tokenizer}

The tokenization procedure must be able to tokenize any text string into tokens that all exist in the vocabulary of the language model. \textit{Byte-pair-encodings} (BPE) \cite{bpe}, and variations of it, is a common way to tokenize text for transformer language models especially for NMT. BPE strikes a balance between word-level and character-level tokenization by using subword-tokenization. It is able to express almost any string, like character-level tokenization, but without needing to treat each character separately which would result in very long sequences. 

For the model, the number of merges was set to have a resulting total vocabulary size of 50,000, which is close in size to the GPT-2 vocabulary. The Byte-level BPE vocabulary was learned on the entire corpus. Additionally, included into this vocabulary are some special tokens which we added: \textit{<sos>, <eos>} for start and end of text tokens, \textit{<unk>} for unknown tokens just in case there is an error, \textit{<special1>, <special2>, <special3>} tokens reserved for possible of later downstream use when fine-tuning the model for a specific task. The special tokens never appear in the pre-training corpus, and \textit{<special1>} is used later on when fine-tuning to generate headlines.

\section{Building a Finnish GPT-2}

Our approach to creating a headline generating model is based on fine-tuning a language model learned by unsupervised generative pre-training. As such a model does not exist for Finnish, we have to train one.

\subsection{Model Specifications}
The language model in this paper are decoder Transformers, with a few key modifications. The modifications closely follow those made to GPT-2 as compared to the original Transformer.

Positioning of the layer normalization has been to follow GPT-2. Originally layer normalization was applied after the residual connections. This was modified by moving layer normalization to the input of each sublayer, and adding an additional layer normalization to the output of the final self-attention layer.

Positional embeddings are learned instead of sinusoidal. The reason for this is that BERT and the GPT variants use learned positional embeddings as well. This involves adding another embedding matrix to the neural architecture in addition to the token embedding matrix. The difference is that the position embedding matrix keys are the position integers of a token relative to the text it resides in, while the token embedding matrix has the vocabulary id of the token as the index.

Again following GPT-2, the network parameters are initialized by sampling from $N(0, \frac{0.02}{\sqrt{n}})$, where $n$ is the number of residual layers. From our experiments, this change is crucial for the convergence of larger model sizes. The model sized discussed in this paper which are of size $L$ and larger did not converge at all without this change.

Like GPT-2, Gaussian Error Linear Unit (GELU) \cite{gelu} was used as the activation function in the network instead of ReLU \cite{relu}. We used the AdamW \cite{adamw} optimizer with a learning rate $\alpha$. For the final model, $d_{model}=1280$ and $n_{warmup} = 2000$. This was done by doing several restarts and observing when the gradients overflow and the training breaks down as the learning rate is increased during warmup. The formula is tuned so that the peak learning rate is lower than the learning rate was when overflow was observed. 

GPT-2 had additional L2 regularization which is omitted in this paper work. Finally, our models do not use dropout regularization. This is due to seeing better validation set convergence without it when testing hyperparameters on smaller test training runs. Since training the Transformer model takes time, omitting dropout regularization allows the model to converge slightly faster in terms of time. We ran several small-scale experiments with varying degrees of dropout and found that it did not significantly affect the end validation perplexity.

\subsection{Results of Pre-training}
\textit{Perplexity} (Equation~\ref{eq:ppl}) is a commonly used measure of the performance of a language model \cite{gpt, gpt2, gpt3}. The perplexity of a model given a text is calculated based on the probability assigned to each actual token in the text given its context. A convenient way to calculate it is by exponentiating the negative log likelihood loss: 

\begin{equation}
\label{eq:ppl}
    ppl = e^{{-\sum_{t=1}^{n}log(p(\vect{x}^{(t)}|\vect{x}^{(1)},\ldots,\vect{x}^{(t-1)},\vect{\theta})}}
\end{equation}

A language model with a vocabulary of size $V$ will have a perplexity score of exactly $V$ if it always predicts the uniform distribution for each token. A random language model will average around $V$ perplexity as well. This is because on average, the correct token in a text is given $\frac{1}{V}$ probability by such a LM. Conversely, a LM that always predicts the token correctly with 100\% assigned probability will have a perplexity score of 1. 

The resulting train and test perplexities achieved for the various models are found in Table~\ref{table:lmresults}. The medium and large models were trained for 4 epochs each, while the second large model was trained for less than 3 epochs. The training times were approximately 2 weeks for all models, and the medium and first large models were trained on half of the full corpus. This test reveals that in this case, the size of the training set has more impact on the resulting model perplexity than the size of the model. The implication is that increasing the size of the model won't increase the quality of the model significantly if the amount of training data is insufficient. Due to this and that it would have taken more than 2 weeks to train the XL sized model, we chose not to train a full XL sized GPT model. Additionally, the generative performance from manual testing was good enough.

\begin{table}
\centering
\begin{tabular}{||c| c | c ||} 
 \hline
 Size & ppl\_train & ppl\_test \\ [0.5ex] 
 \hline\hline
 medium & 17.9 & 22.9\\
 \hline
 large & 17.4 & 21.6\\
 \hline
  large2 & 14.0 & 17.8\\
 \hline
\end{tabular}
\caption{Train and test perplexity scores.}
\label{table:lmresults}
\end{table}

\section{Headline Generation}

This chapter describes how we tackle the task of conditional headline generation by using transfer learning. The final Finnish LM is used as the base model. we describe the training procedure and decoding algorithm design first, followed by the description of a domain-expert evaluation of the headline generative performance.

The final Finnish LM is loaded from its latest checkpoint, including both its parameters and optimizer state. Training is resumed but with changes to the learning rate, input structure, and validation generation. The loss calculation is altered as well, to focus the learning purely on the task of headline generation.

The training, validation and testing corpora are filtered, removing texts that do not have a headline or are not news articles. The texts are re-formatted, clipping the body of the text to the first 448 tokens. The special token \textit{<special1>}, not previously shown to the model, is appended to the end of the clipped body text. The headline of the text is then appended following the special token, followed by the \textit{<eos>} token signifying the end of the output. The idea is to learn a pattern where text that follows a special token is always a headline summarizing the preceding text, followed by the end token. Ideally then when the LM is given any text prompt ending in the special token, the output of applying a generation algorithm would be a relevant headline. The clipping procedure results in a portion of the news articles body text not to be completely shown to the model. This clipping must be done due to the model having a maximum context width of 512 tokens in order to fit the headlines at the end. We chose to keep the beginning of the texts due to news articles being structured in a way where the most important content tends to be written in the beginning of the article. 

The data is no longer fed to the model in a dense square matrix format as when pre-training the LM. That method would be separating the headline of an article and the tokens of the article itself much of the time into separate rows in the input tensors. Ideally the corpus would be sorted according to length and fed to the model in variable sized batches of instances with similar length. We omitted this step for convenience, as we only ran the fine-tuning runs for one epoch each.

Sampling based algorithms such as nucleus sampling don't seem to lend themselves well for the task of headline generation. This can be seen by simply observing the random nature of the headline generation when using sampling based methods, both from the validation output and from manual testing. The requirements for headlines are stricter than that of creative text continuation, in that the headline must at least summarize accurately the article, and not invent things not stated in the text.

Beam Search works better for this task. The problem with Beam Search for practical applications though, is that if you want several headlines, it produces the same headline with only slight variations. Often, with just one word differences at the end. 

Diverse Beam Search (DBS) \cite{dbs} is an alternative to BS which addresses the diversity issue by decoding in a doubly greedy manner, optimizing both the sequence probability under the model as well as the diversity.

At a high level, the \textit{B} beams are divided into \textit{G} groups. In addition, a similarity penalty is introduced. This penalizes subsequent groups sequence probability score by a similarity penalty term multiplied by $\lambda$, a parameter for similarity penalty strength. In this work, the similarity score is the integer number of times in previous beam groups the proposed token has been selected during this step. 

In this algorithm we have \textit{G} separate groups of beam-search. For each decoding step, the groups of beam-search are advanced in consecutive order. For each consecutive token in the decoding process, the first group has no diversity penalty from previous groups, and as such is simply beam search. For each consecutive group, regular beam search is conducted but with the sequences penalized when the proposed token has been used in previous groups during this step. This adds diversity between the groups already from the first token if $\lambda$ is high enough, due to each group beginning the headline with a different token.

Vanilla DBS does not address the repetitiveness problem. While repetitiveness seems to happen less when generating headlines, it still does happen that a name or sentence is repeated. For this reason, we added a second penalty which is beam-specific: $\lambda_{repeat}$. This is a penalty applied to the probability score of continuations to a sequence when the proposed token has previously been used in the sequence in question. This is mainly to prevent outputs such as \textit{"Niinistö tapasi Niinistön"} (Niinistö met Niinistö). If $\lambda_{repeat}$ is set too high, then grammar can suffer due to proper suffixes being penalized too harshly. One could say that if the model was good, this should be unnecessary. Unfortunately, it seems that this repetitive behaviour is common in this type of MLE language model optimization.

The likelihood under the model for a sequence in Beam Search in general is the joint probability of the sequence calculated using the Chain Rule of Probabilities by multiplying each token probability conditioned on the context together. In this case of generating headlines, this causes shorter headlines to have a higher probability. They are usually safer but more boring headlines. In order to combat this, we added a decay parameter $\beta$. This parameter is multiplied together with the current log-probability of the sequence so far before adding the log-probability of the proposed token to it. The result is equivalent for $\beta = 1$ and results in longer headlines when $\beta < 1$.

Our implementation of DBS has 6 parameters in total. \textit{G} and \textit{B} for the number of groups and number of beams per group. We selected $G=4$ and $B=2$ for 2 beams per group and 4 output headlines, totalling in 8 beams. The maximum length of a headline is another hyperparameter which we set to 48 tokens. The 3 remaining hyperparameters $\lambda$, $\lambda_{repeat}$ and $\beta$ were tuned algorithmically, because it was too much manual work with unclear results to tune these manually. 

We used Gaussian Process optimization \cite{gpopt} to select these 3 parameters. The objective function we used was the BLEU \cite{bleu} score of the generated set of headlines with regards to the true headline for 100 articles. We opted for GP optimization instead of grid search as grid search would have taken too long, as generating one set of headlines once already takes several seconds.

For $\lambda$ we observed two separate points of interest where the target (BLEU) is at maxima. These are consequently the points with highest search density for this hyper-parameter. The final values for the hyper-parameters were $\lambda = 0.71$, $\lambda_{repeat} = 3$ and $\beta = 0.87$. Notably, 3 was the maximum we had set for $\lambda_{repeat}$, making higher values possibly better still.

\section{Results and Evaluation}

As previously mentioned, the evaluation of models should be conducted in a way that measures the performance of the actual desired task at hand. For this reason, calculating BLEU or similar on an offline corpus is not an accurate representation of the performance of the model when it comes to generating real world headlines. The question we seek to answer in this paper is how well can this model perform in real-world use in the newsroom as a tool to help editors headline articles. 

\subsection{Study Design}

To thoroughly answer the research question, we generated a set of headlines for new articles, and had domain-experts evaluate them by hand according to three key criteria. We picked 100 random news articles from Helsingin Sanomat (HS) and Ilta-Sanomat (IS) not contained in the original corpus. For each article, we generated four headlines using our implementation of DBS and the optimized parameter set. We made an Excel worksheet\footnote{https://zenodo.org/record/5985728} where each article had its text in one column, and in another column its four generated headlines as well as the real original headline in a random position in the headline set. The worksheet has a column for each of the three criteria which the evaluators fill with 1 for the headline passing the criterion and 0 otherwise. The criteria are in order of difficulty for the model to achieve, with the first criterion being the easiest and the third criterion being the most difficult. Additionally, passing a criterion means passing the preceding criteria as well. The criteria are language, usable and good.

\textbf{Language} If disregarding the article text, is the headline on its own correct Finnish? Does this headline make sense to a human being? We elected to have this criterion separate from the next one to get a better understanding of where and how the performance breaks down.

\textbf{Usable} Could this headline be used for the given text in the real world? Does it match the text in the news article without misquoting and without errors? This is the most important question in terms of how good this model is for real-world use.

\textbf{Good} Is this headline good enough for the editor to be comfortable publishing the article with it without feeling the need to edit it or come up with variants? This final criterion is a subjective one but we decided to keep it separate from the usable criterion as they are fundamentally different.

Additionally, there's an optional open feedback column, as well as summary open feedback at the end of filling in the excel.

Three editors, one from Helsingin Sanomat and two from Ilta-Sanomat volunteered to perform this evaluation. Each one has extensive experience in headlining articles, sometimes coming up with dozens of headlines in a day. The final answers for each question are selected as the majority vote of the three. It took two weeks for them to fill in their answers. The real headline was inserted randomly as a control for possible anti-machine bias and as a baseline reference \cite{framing}.

Out of the 500 headlines, 467 received an evaluation from all three evaluators. Some of the headlines were not evaluated due to the source text having been incorrectly parsed, leaving out names of people and places and was deemed by the evaluator(s) to be best left unanswered. Some headlines seem to have been simply forgotten. Most of the following tables have the metrics for the real and the generated headlines separate for baseline reference.

The acceptance percentages for each of the three evaluation criteria per individual evaluator are shown in Table~\ref{table:evaluatorstats}. We can see that evaluator A seems to have been able to distinguish between the real and generated headlines better than the other two evaluators, while evaluator B was the most forgiving.

\begin{table}[h!]
\centering
\begin{tabular}{||c| c |c |c |}
\multicolumn{4}{c}{Language} \\
 \hline
 Evaluator & A & B & C  \\ [0.15ex] 
 \hline\hline
 Real & 1.0 & 0.97 & 0.785 \\
 \hline
 Generated & 0.79 & 0.90 & 0.775 \\ [0.15ex] 
 \hline
\end{tabular}
\centering
\begin{tabular}{| c |c |c |} 
\multicolumn{3}{c}{Usable} \\
 \hline
 A & B & C  \\ [0.15ex] 
 \hline\hline
 0.91 & 0.80 & 0.77 \\
 \hline
 0.22 & 0.43 & 0.37 \\ [0.15ex] 
 \hline
\end{tabular}
\centering
\begin{tabular}{| c |c |c ||} 
\multicolumn{3}{c}{Good} \\
 \hline
  A & B & C  \\ [0.15ex] 
 \hline\hline
 0.84 & 0.76 & 0.47 \\
 \hline
 0.13 & 0.40 & 0.20 \\ [0.15ex] 
 \hline
\end{tabular}
\caption{The response acceptance ratio for each evaluator separately for Language, Usable and Good criteria separated by real headlines and generated headlines.}
\label{table:evaluatorstats}
\end{table}

The inter-annotator agreement per criterion measured by Fleiss' kappa~\cite{fleiss} is shown in Table~\ref{table:agreement}. Fleiss' kappa represents the degree of agreement when accounting for agreement by chance based on the ratio of passing versus rejecting the criteria. A positive number between 0 and 1 means there is more agreement than by chance, while a negative number between 0 and -1 indicates more disagreement than by chance. 

For real headlines the degree of agreement is negative and close to chance. This is expected, as the majority of real headlines pass the criteria and the criteria are inherently slightly subjective. 

The agreement in the three criteria for the generated headlines was modest but clearly greater than chance. The merely modest inter-annotator agreement shows numerically how the generated headlines often have errors that are hard to detect, as clearer errors would yield a high degree of agreement. The goodness criterion has the lowest inter-annotator agreement despite the model failing this criterion the most, as it is the most subjective.
\begin{table}[h!]
\centering
\begin{tabular}{||c| c |c |c ||} 
 \hline
 Type & Language & Usable & Good  \\ [0.5ex] 
 \hline\hline
 Real & -0.09 & -0.02 & -0.07 \\
 \hline
 Generated & 0.35 & 0.38 & 0.30 \\ [1ex] 
 \hline
\end{tabular}
\caption{Inter-annotator agreement measured by Fleiss' kappa.}
\label{table:agreement}
\end{table}

The headlines performed equally well per brand, as seen in Table~\ref{table:product}. The language criterion was the only criterion where there was a notable difference between the brands. The model seems to have a slightly easier time with HS articles. 

\begin{table}[h!]
\centering
\begin{tabular}{||c| c |c |c ||} 
 \hline
 Brand & Language & Usable & Good  \\ [0.5ex] 
 \hline\hline
 HS & 0.91 & 0.31 & 0.20 \\
 \hline
 IS & 0.82 & 0.30 & 0.21 \\ [1ex] 
 \hline
\end{tabular}
\caption{Acceptance rates by brand. Both brands had approximately the same amount of headlines.}
\label{table:product}
\end{table}

The final result of the survey where the headlines are scored for each criteria according to a majority vote is shown in Table~\ref{table:summarystats}. A headline passes a criterion if at least two out of the three evaluators vote to pass. we have the real control headlines separate from the generated headlines as a baseline reference. Additionally, these tables show both the total acceptance rates as well as acceptance rates for headlines that have passed the preceding criteria. We can see that the language criterion is where the model performs by far the best as expected. The performance breakdown is clearly between the language and the usable criteria, as only 35\% of headlines that pass the language criterion pass the usable criterion as well. Of those that do however, 68\% pass the difficult final criterion.

\begin{table}[h!]
\centering
\begin{tabular}{||c| c |c |c ||} 
 \hline
 Type & Language & Usable & Good  \\ [0.5ex] 
 \hline\hline
 Real & 1.0 & 0.89 & 0.89 \\
 \hline
 Generated & 0.87 & 0.35 & 0.68 \\ [0.5ex] 
 \hline
\end{tabular}
%\caption{Summary for generated versus real headlines majority vote responses for headlines that passed the preceding criteria.}
%\label{table:summarystatscarryover}
%\end{table}
%\begin{table}[h!]
\begin{tabular}{||c |c |c ||} 
 \hline
 Language & Usable & Good  \\ [0.5ex] 
 \hline\hline
 1.0 & 0.89 & 0.79 \\
 \hline
 0.87 & 0.31 & 0.21 \\ [0.5ex] 
 \hline
\end{tabular}
\caption{Summary for generated versus real headlines majority vote responses. The first table shows metrics for headlines that have passed the preceding criteria, while the second table shows the total for all headlines.}
\label{table:summarystats}
\end{table}

\section{Discussions}

Although free text generation is not the focus of this paper, the generative capabilities of the Finnish GPT are still noteworthy and relevant for the headline generation task. Evaluating the generative performance of a language model in-depth is a very time-consuming task, and we will outline the major findings we have with this particular model here. These findings mostly come from manually giving the model different prompts and using different parameterizations of top-p and top-k sampling to generate continuations. 

From the logging of validation top-k next tokens and their assigned probabilities during training, it is clear that the output probability distribution for the next token becomes sharper as the training run progresses. The shape of the output distribution has a significant impact on sampling based decoding output, as sharp distributions produce less varied output. This makes generating a snippet during validation by using a fixed set of parameters for the sampling algorithm a poor way of gauging the progression of the true generative capability of a language model. Note that the temperature parameter directly affects the sharpness of the output distribution as well. For both top-k and top-p sampling, we found that a range of 0.6-1.0 was the usable range for temperature. Values of over 1 result in very random and nonsensical text, while values of less than 0.6 became very repetitive.

Repetition is known as the most prevalent pathology in text generation using deep neural language models \cite{repetition}. This pathology occurs the worst the greedier the decoding algorithm. Greedy decoding and vanilla beam-search decoding which try to find the approximate MLE generation suffer from this the most. Top-k and top-p sampling partially combat this, by using the random nature of the sampling to break repetition loops. The true reason for the repetitive behavior of current language modeling solutions is not understood. 

The first form of repetition is in the form of repeating entire or partial sentences one, several or even infinite times, sometimes with a slight variation. This makes for text that does not resemble human text, and is not desireable. 

The second form of repetition is the repetition of names, places and objects in a way that does not semantically make sense. An engineered example: \textit{"Sauli Niinistö tapasi keskiviikkona Tasavallan Presidentti Sauli Niinistön"} (Sauli Niinistö met the President of the Republic Sauli Niinistö on Wednesday). This sentence does not make sense, as a person cannot meet himself. In this case, it seems that the locally highly correlated continuation to "Tasavallan Presidentti" (President of the Republic), which is "Tasavallan Presidentti Sauli Niinistö" (the President of the Republic Sauli Niinistö) in the training data, overrides the fact that he should never be the prediction in this context conditioned on him being already mentioned in the sentence.

The opposite of repetition can occur. It can occur with greedier decoding as well but is more pronounced with sampling based decoding. Again, we class these into two main categories. 

The first category is the direct opposite of the repetition of names and places. This is when a text mentions the name of a person, and the generated output suddenly swaps out the name for another name and continues the text with the new name. The severity of this varies depending on prompt length and context. If the context is very U.S Presidential heavy and the name supplied is \textit{Donald Bump}, it will likely be "corrected" to \textit{Donald Trump} due to the sheer volume of support for the latter in the corpus.

Interestingly, the second form of correction may actually have some use. This is when a sentence is repeated, but with more probable grammar. As an example, there may be a grammatical error in a quote, the model can then accidentally correct the grammatical error when repeating the quote.

\section{Conclusions}

The task was to create and evaluate a headline generation algorithm in the context of helping editors in the newsroom in the creative process. This is what was done in this paper. A neural language model was pre-trained on Finnish text, and fine-tuned to generate headlines. A decoding algorithm for diverse output was implemented. The resulting generated headlines were evaluated by domain experts to gauge the feasibility of this model in actual use. This sort of evaluation is the first we've seen when it comes to evaluating a headline generation algorithm.

The final conclusions are that while most of the time the generated headlines are very close to being usable, this particular implementation is far from ready in any sort of automated system. This comes as no surprise, as even with near perfect usability performance it would still not be used without a human in the loop. The algorithm in this work has potential and an expressed interest as a creative aid for the headlining process. 

The most common errors especially for the language and usable criteria are clear and have potential solutions. Some of the errors can be tackled by pre- and post-processing such as the unsightly special character code printouts. The repetition errors, which were the majority of language errors, can be reduced with the repetition penalty. We hypothesize that several of the errors could be tackled with an adversarial and/or active reinforcement learning approach. The problem with generative pre-training seems to be that the model is only trained with what is correct, with everything else being equally incorrect. In reality when producing headlines, this is not the case.

The next steps would be the low-hanging fruit: tackling the error types specifically with parsing fixes and repetition penalty, as well as letting the fine-tuning process converge more. After that, trying more strongly correlated metrics as the decoding algorithm base score, and trying encoder-decoder type approaches as well as active reinforcement learning or adversarial approaches.

\section*{Acknowledgments}
Special thanks to Pipsa Havula (IS), Esa Mäkinen (HS) and Simo Holopainen (IS) from Helsingin Sanomat and Ilta-Sanomat for making the effort to evaluate the headline worksheet and provide feedback. Thanks to Helsingin Sanomat and Ilta-Sanomat for the corpus of news articles which constituted the bulk of the data used. Additionally we wish to thank the Finnish Computing Competence Infrastructure (FCCI) for supporting this project with computational and data storage resources. This work was partially financed by the Society of Swedish Literature in Finland with funding from Enhancing Conversational AI with Computational Creativity, and by the Ella and Georg Ehrnrooth Foundation for Modelling Conversational Artificial Intelligence with Intent and Creativity. This research has received mobility funding from Nokia Foundation under grant number 20220193.

\bibliography{anthology,custom}

\end{document}